\documentclass[10pt,twocolumn,letterpaper]{article}
 
\usepackage{kotex}
\usepackage{style}
\usepackage{times}
\usepackage{epsfig}
\usepackage{graphicx}
\usepackage{amsmath}
\usepackage{amssymb}
\usepackage{booktabs}
\usepackage{makecell}
\usepackage{float}
\usepackage{caption}
\usepackage{xcolor}
\definecolor{DarkGreen}{rgb}{0.0, 0.7, 0.0}
\definecolor{DarkBlue}{rgb}{0.0, 0.0, 1.0}
\definecolor{SoftGreen}{RGB}{143,188,143}
\usepackage{float}
\usepackage{makecell}
\usepackage{subcaption}
\usepackage{graphicx}
\usepackage{hyperref}

\newcommand{\tb}[1]{\textcolor{blue}{#1}}

\newcommand{\tr}[1]{\textcolor{red}{#1}}
\newcommand{\tR}[1]{\textcolor{red}{\textbf{#1}}}

\begin{document}

\title{PointCubeNet: 3D Part-level Reasoning with 3x3x3 Point Cloud Blocks}

\author{
Da-Yeong Kim and Yeong-Jun Cho\textsuperscript{*}\\
Department of Artificial Intelligence Convergence\\
Chonnam National University, Gwangju 61186, South Korea\\
{\tt\small \{qetuo090909, yj.cho\}@jnu.ac.kr}
}

\twocolumn[{ 
\maketitle
\centering
\vspace{-0.2cm}  
\includegraphics[width=1\textwidth]{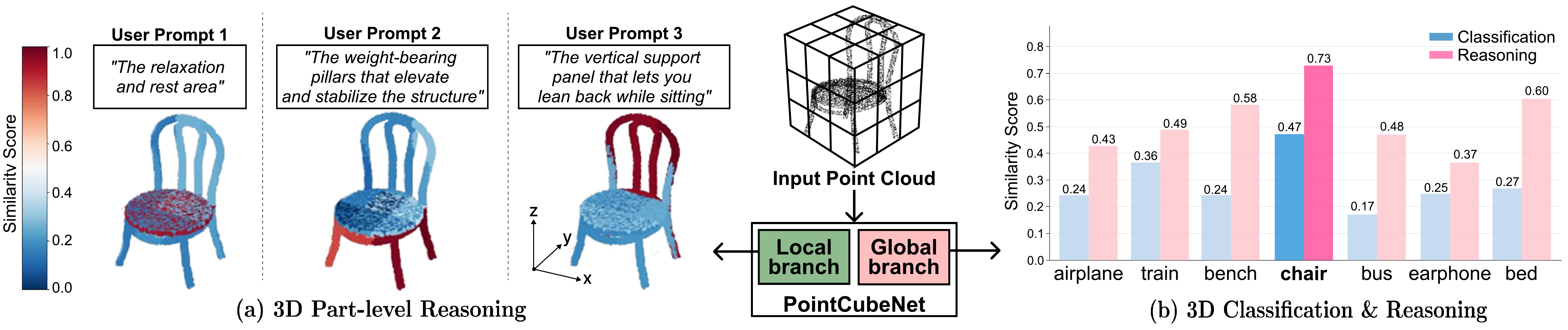} 
\captionof{figure}{
   \textbf{Overview and potential applications of the proposed PointCubeNet.} This method analyzes the global and local features of the point cloud in 3×3×3 point cloud blocks. By leveraging this capability, it can understand both an entire 3D object (e.g., classification and reasoning) and its local parts in an unsupervised manner.
}
\vspace{0.6cm} 
\label{fig:teaser}
}]

\maketitle

\begin{abstract}   
   In this paper, we propose a novel multi-modal 3D understanding framework called PointCubeNet, which performs 3D classification, reasoning, and part-level reasoning.
   PointCubeNet comprises global and local branches.
   The proposed local branch, structured into 3x3x3 local blocks, enables part-level analysis of point cloud sub-regions with the corresponding local text labels. 
   Leveraging the proposed pseudo-labeling method and local loss function, PointCubeNet is effectively trained in an unsupervised manner. 
   The experimental results demonstrate that understanding 3D object parts enhances the understanding of the overall 3D object. 
   In addition, this is the first attempt to perform unsupervised 3D part-level reasoning and achieves reliable and meaningful results. 
   The code for PointCubeNet is available at \url{https://url}.
\end{abstract}
\vspace{-10pt}

\section{Introduction}

\begin{figure}
   \centering
   \begin{subfigure}[b]{0.42\columnwidth}
     \includegraphics[width=\linewidth]{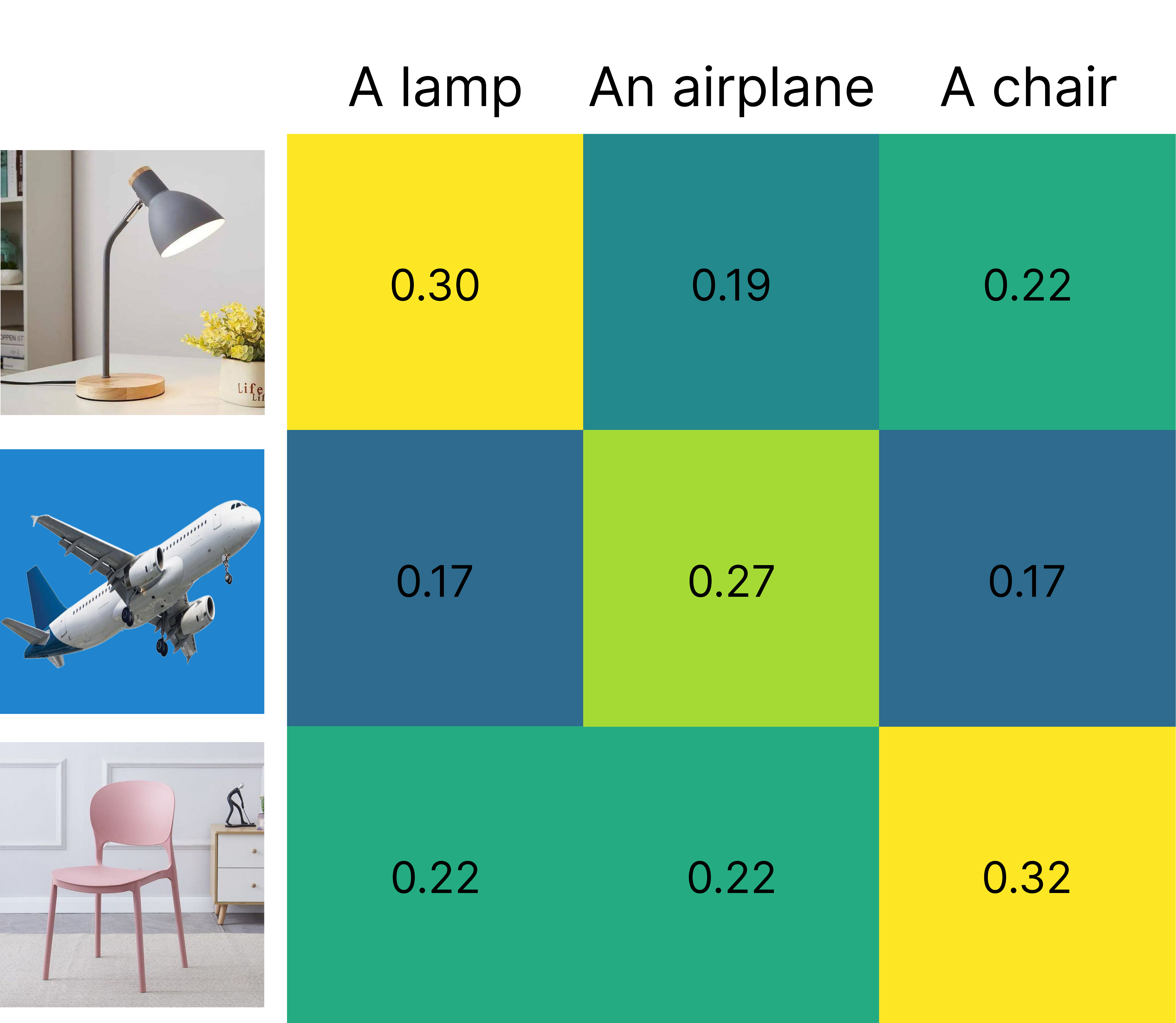}
     \caption{Classification}
   \end{subfigure}
   \hspace{15pt}
   \begin{subfigure}[b]{0.41\columnwidth}
     \includegraphics[width=\linewidth]{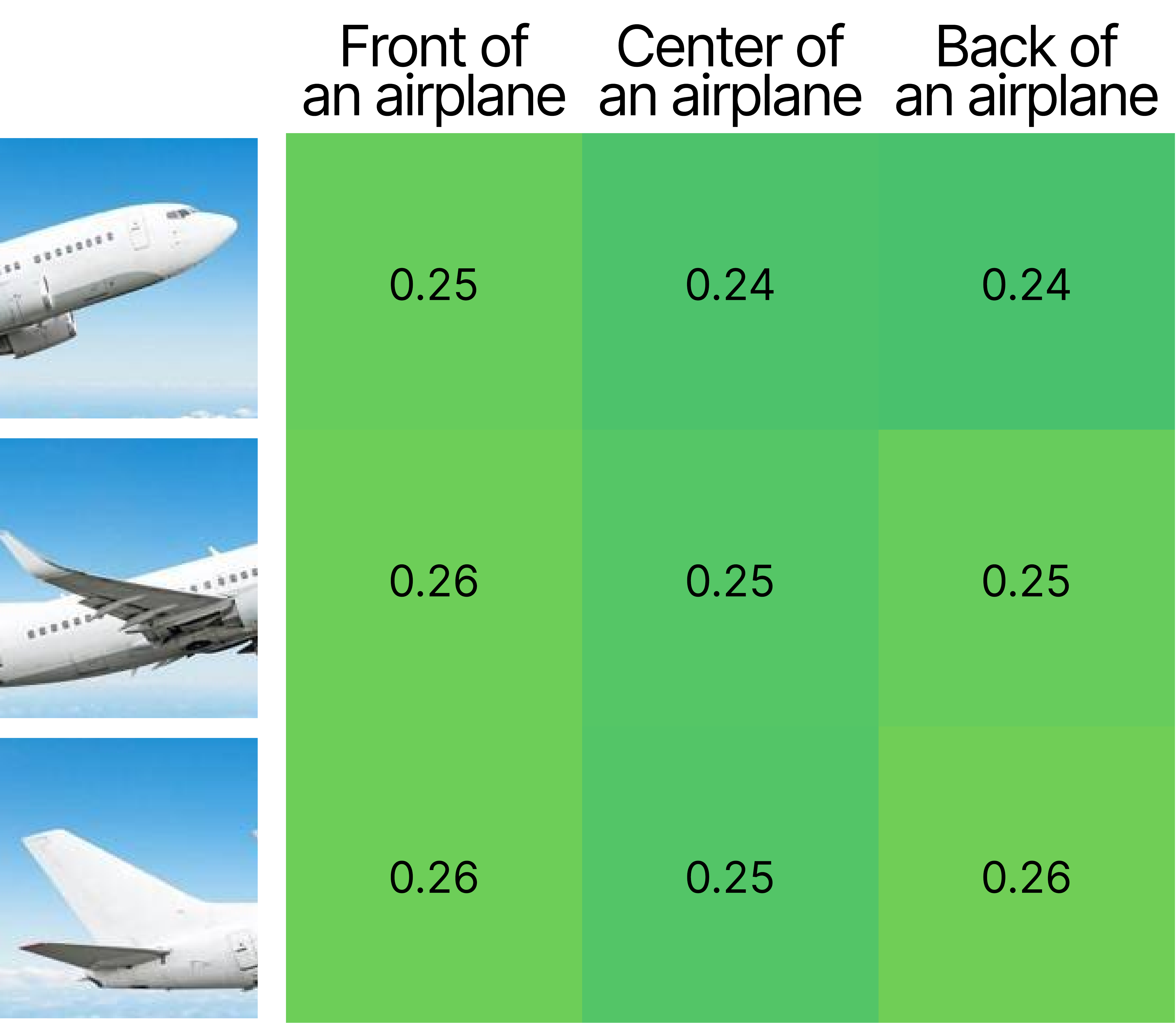}
     \caption{Part classification}
   \end{subfigure}
   \caption{\textbf{Classification results of CLIP~\cite{CLIP}.} It understands the entire object well but fails to distinguish its parts.}
   \label{fig:clip_part}
   \vspace{-15pt}
 \end{figure}

Advancements in deep learning have significantly improved three-dimensional (3D) object understanding tasks~\cite{PointNet, qi2016volumetric, su2015multi} such as 3D detection, classification, and part segmentation. 
Recently, beyond the single-modality process, many studies have explored multi-modal 3D understanding by integrating vision-language models (VLMs) that learn joint representations of visual and textual data. 
Most approaches~\cite{CLIP2Point, xue2024ulip, PointCLIPV2} have employed CLIP~\cite{CLIP} as their VLM, leveraging its strong generalization and zero-shot capabilities.
However, these methods heavily depend on the structure and performance of VLMs.
Most VLMs were not initially designed for 3D understanding; hence, their image encoders necessitate projecting 3D point clouds onto  multi-view 2D images.
This 3D to 2D projection can result in losing critical 3D structural and fine-grained details.
In addition, although many VLMs effectively capture global relationships between visual and textual information, they struggle to capture local and fine-grained relationships, as presented in Fig.~\ref{fig:clip_part}.
This limitation arises from the contrastive learning process, which focuses solely on global correspondence without modeling local relationships.
Thus, previous multi-modal 3D understanding methods cannot handle part understanding tasks.

In this paper, we propose PointCubeNet, a multimodal 3D understanding framework that performs 3D classification and reasoning while enabling 3D part-level reasoning. 
PointCubeNet is a model capable of analyzing the entire structure and the individual parts of an input 3D object. 
As illustrated in Fig.~\ref{fig:teaser}, the reasoning task aims to identify objects or their parts by aligning visual and textual features, even from indirect text descriptions.
PointCubeNet directly analyzes raw 3D data using a point-based 3D encoder~\cite{DGCNN, PointNet} without requiring voxelization, 2D projection, or reliance on the 2D image and text encoders of CLIP.

PointCubeNet consists of global and local branches that effectively associate text labels with global and local 3D visual features. 
In particular, the proposed local branch is designed to analyze sub-regions of point clouds, which are uniformly grouped into 3×3×3 local blocks.
The critical challenges in training PointCubeNet include generating large-scale text annotations for local object parts and designing an effective local loss function to optimize the local branch.
To address this aim, we propose a pseudo-labeling method for local text annotation, enabling the model to learn visual-textual relationships for local structures in an unsupervised manner. 
Next, we propose a novel similarity-based local loss that effectively aligns 3D local features with the corresponding local text labels.

To validate the proposed methods, we employed two 3D benchmark datasets, \texttt{ModelNet}~\cite{ModelNet} and \texttt{ShapeNet}~\cite{ShapeNet}.
PointCubeNet achieves superior performance in both 3D classification and reasoning tasks.
In addition, explicit experiments demonstrate that understanding parts of 3D objects by the proposed local branch in PointCubeNet improves the understanding of whole 3D objects.
Furthermore, PointCubeNet can also perform part-level reasoning without manual text annotations for each part.
As shown in Fig.\ref{fig:part-level_reasoning} and Fig.\ref{fig:zero-shot_part-level_reasoning}, it produces reliable and meaningful part-level reasoning results.

The primary contributions of this work are as follows:

\noindent $\bullet$ This is the first attempt to process 3×3×3 local blocks to achieve unsupervised 3D part-level reasoning.

\noindent $\bullet$ We propose a novel multi-modal 3D understanding framework that directly analyzes raw 3D point cloud data without relying on pretrained VLMs such as CLIP encoders.

\noindent $\bullet$ We validate that understanding parts of 3D objects enhances the understanding of the whole 3D object.

\noindent $\bullet$ The experimental results demonstrate the effectiveness of the proposed methods across various scenarios.


\section{Related Work}
\noindent \textbf{3D Point Cloud Understanding.} \quad
Recent advancements in deep learning have significantly improved the performance of 3D object understanding tasks, such as detection, classification, and segmentation.
Point-based approaches~\cite{li2018pointcnn, DGCNN, PointNet, PointNet2} aim to analyze raw point clouds directly using neural networks. 
PointNet~\cite{PointNet} employs max-pooling to extract permutation-invariant global features via simple multi-layer perceptrons.
However, this method struggles to capture fine-grained local geometric structures.
To address this limitation, PointNet++\cite{PointNet2} introduces hierarchical grouping to capture local features more effectively, whereas the dynamic graph convolutional neural network (DGCNN)~\cite{DGCNN} models local relationships. 
These approaches underscore the need to extract global and local features for improved 3D object representation.

Inspired by the success of CNNs~\cite{he2016deep, szegedy2015going} in 2D image understanding, many methods have been proposed to process 3D point clouds using voxel-based 3D CNNs~\cite{atzmon2018point, maturana2015voxnet, ModelNet} or by projecting point clouds onto 2D multiview images to apply existing 2D CNNs~\cite{qi2016volumetric, su2015multi}, rather than use raw point clouds. 
These methods employ the robust performance of well-trained CNNs, achieving effective results on 3D tasks. 
However, transformations, such as voxelization or multiview projections, may lead to losing critical 3D information and fine-grained details.

\noindent \textbf{Vision-Language Models.} \quad Recent research has expanded beyond understanding a single modality to training Vision-Language Models (VLMs)~\cite{jia2021scaling,li2022blip, CLIP} that learn joint representations of visual and textual data.
Notably, CLIP~\cite{CLIP} learns a shared embedding space for images and text, enabling a wide range of applications.
It was trained on large-scale datasets (e.g., ImageNet\cite{ImageNet} and LAION-5B~\cite{schuhmann2022laion}), exhibiting strong generalizability while supporting zero-shot and few-shot learning.

\begin{figure*} [t]
    \centering
    \includegraphics[width=1\textwidth]{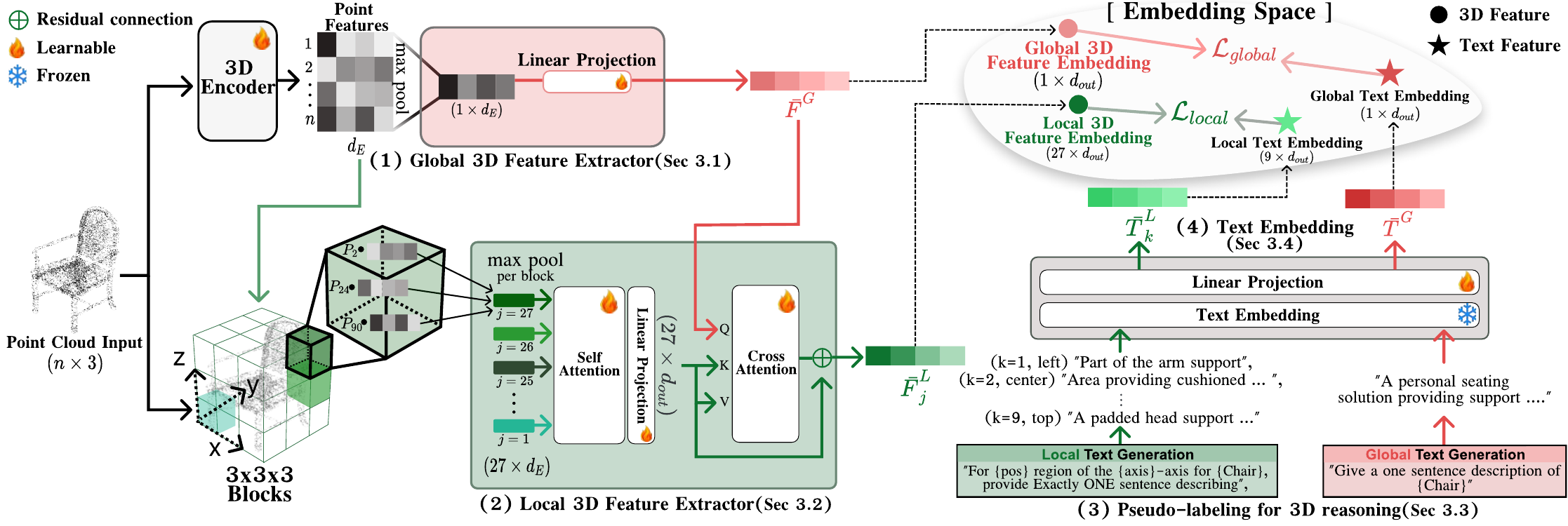}
    \caption{\textbf{The pipeline of PointCubeNet.} It consists of a \colorbox{pink}{global branch} and a \colorbox{SoftGreen}{local branch}, with four main components: (1) a global 3D feature extractor, (2) a local 3D feature extractor, (3) global and local pseudo-labeling for 3D reasoning, and (4) a text embedding module. The network is trained to measure the similarity between global and local 3D feature embeddings and text embeddings. It performs 3D classification, reasoning, and part-level reasoning without training any head structures.}
    \label{fig:proposed_framework}
\end{figure*}

\noindent \textbf{Multi-modal 3D Understanding.} \quad  
The field of 3D understanding has increasingly integrated multi-modal approaches.
Most studies~\cite{CLIP2Point, ULIP, xue2024ulip, PointCLIP, PointCLIPV2} on multi-modal 3D understanding have applied the visual and text encoders from the well-trained CLIP~\cite{CLIP}.
These methods rely on the robust performance and zero-shot classification capabilities of CLIP and heavily rely on its abilities and limitations. 
For instance, these approaches require the 2D projection of point clouds, losing critical 3D structural information because CLIP is designed to process only 2D images. 
Fig.~\ref{fig:clip_part} reveals that CLIP struggles with fine-grained local part inference in images. 
Hence, CLIP-based 3D understanding methods~\cite{CLIP2Point, xue2024ulip, PointCLIPV2} are limited to object classification and struggle with tasks requiring detailed 3D part understanding, such as part segmentation and part-level reasoning. 
A recent study~\cite{PARIS3D} proposed reasoning-based 3D part segmentation via another VLM (i.e., LISA~\cite{LISA}); however, it requires point cloud-level annotations, which are  labor-intensive and limit scalability.

\section{Proposed Methods}
\label{sec:proposed}
PointCubeNet is a multi-modal 3D understanding framework that understands whole 3D objects and their parts. 
Thus, two branches are designed in PointCubeNet: global and local as shown in Fig.~\ref{fig:proposed_framework}.
\subsection{Global 3D Feature Extractor}
\label{subsec:global}

A 3D object $\mathbf{O}$, represented by a set of points~(i.e., a point cloud) is defined as
$\mathbf{O} = \{P_i \mid i=1,..., n\}$, where $P_i$ denotes the $i$th 3D point, and $n$ represents the total number of points.
Each point $P_i\in\mathbb{R}^{3}$ represents a 3D coordinate $\left(x,y,z\right)$. 
This work does not utilize other channels of point clouds such as color and normal.
Thus, the 3D object is represented by $n\times 3$ values.
To extract 3D features from point clouds, we utilize a pre-trained 3D encoder $E_{P}$ by
\begin{equation}
   [f_1, f_2, ..., f_n] = E_{P}\left([P_1, P_2, ..., P_{n}]\right),
\label{eq:feature_ext}
\end{equation}
where $E_P: \mathbb{R}^3\rightarrow \mathbb{R}^{d_{E}}$, and $d_E$ denotes the encoded feature dimension.
Any 3D encoder~\cite{DGCNN, PointNet, PointNet2} can be employed as the baseline of this framework. 
The encoder obtains $n$ 3D features $f_i\in \mathbb{R}^{d_{E}}$ with the potential to support further 3D understanding tasks.

Inspired by~\cite{DGCNN, PointNet}, we apply max pooling along the first dimension~($n$) of the 3D features as follows
\begin{equation}
    F^{G} = \text{maxpooling}\left([f_1, f_2, ..., f_n]\right).
\end{equation}
This pooling layer ensures permutation invariance and enables global feature extraction. 
Pooling obtains a global 3D feature $F^G \in \mathbb{R}^{1 \times d_E}$ for the object $\mathbf{O}$.
To improve the representation of the extracted features and align them with text embeddings of $d_{out}$ dimensions as in Sec.~\ref{subsec:text_embedding}, we add a linear projection layer $\mathbf{W}^G\in \mathbb{R}^{d_{E} \times d_{out}}$. 
Then, the global 3D feature embedding is defined as follows: 
\begin{equation}
    \bar{F}^{G}= F^G \cdot \mathbf{W}^G \in \mathbb{R}^{1 \times d_{out}}.
\label{eq:global_proj}
\end{equation}
The global 3D feature embedding $\bar{F}^{G}$ captures the overall properties of the input 3D object via a well-designed 3D encoder $E_{P}$, aligning with the $d_{out}$ dimensionality.

\subsection{Local 3D Feature Extractor}
\label{subsec:local}
The local 3D feature extractor is designed to understand the individual parts of an object.
Previous studies~\cite{PointNet2, PointBERT} have applied the $k$-nearest neighbor approach to divide the entire point cloud of an object into multiple local groups. 
However, these approaches result in dynamic group locations for each object because the positions of $k$ are determined by its local shape and density.
Instead, the 3D points $P_i$ are grouped into $3\times3\times3$ blocks. 
Even if a 3D object does not fit perfectly into a cuboid shape, the $3\times3\times3$ blocks can be adjusted to align with the object. The block approach is simple but effectively enables matching 3D local features to local text embeddings.

After grouping of the point clouds into 27 local blocks, each $j$th block contains multiple 3D points $P_i$ and their corresponding features $f_i$ by $\{P_i \mapsto f_i\} \in \text{block}_{j}$, where $j$ represents the index of the local block and $i$ denotes the index of a point belonging to the $j$th block. 
In this process, $f_i$ is not re-extracted through the 3D encoder $E_P$; instead, it refers to the pre-extracted $f_i$ in Eq.~\ref{eq:feature_ext} corresponding to each $P_i$ in its block.
Max-pooling is performed to extract a representative feature for each block, as follows:
\begin{equation}
    F^{L}_{j} = \text{maxpooling}\left(\{f_i \mid P_i \in \text{block}_j\}\right).
\label{eq:no_attns}
\end{equation}
Then, $\mathcal{F}^{L} = [F^{L}_{1},...,F^{L}_{27}]$ defines a set of local 3D features. 
Each local feature $F^{L}_{j}$ can be directly used after linear projections, similar to the 3D global embedding $\bar{F}^{G}$ in Eq.\ref{eq:global_proj}.   
However, although global embedding is well-learned to represent the entire 3D object, local features lack sufficient analyses and representations for parts of the 3D object. 
To address this problem, we stack additional layers using the vision transformer~\cite{dosovitskiy2021an} to perform self-attention between local 3D features. 
A set of local 3D features $\mathcal{F}^{L}$ is input into the self-attention layers as tokens using
 $\mathcal{F}^{self} = \text{selfAttn}\left(\mathcal{F}^{L}\right)$.
The output clarifies the relationships and dependencies between 3D local features while preserving the input token dimensions $d_E$.
A linear projection layer $\mathbf{W}^L\in \mathbb{R}^{d_{E} \times d_{out}}$ is added for local 3D features by $\mathcal{\bar{F}}^{self} = \{F^{self}_i\cdot\mathbf{W}^{L} \mid F^{self}_i\in \mathcal{F}^{self}\}$ to align those with text embeddings.

Local embeddings can capture relationships between the whole and its parts via cross-attention between the local and global 3D embeddings, resulting in more representative features. 
Therefore, we applied cross-attention between those embeddings as follows:
\begin{equation}
\small
\mathcal{\bar{F}}^{cross}  = \text{crossAttn}(Q=\bar{F}^{G}, K=\mathcal{\bar{F}}^{self}, V=\mathcal{\bar{F}}^{self}),
\end{equation}
where $Q$, $K$, and $V$ denote the query, key, and value, respectively, of the multi-head attention.
Finally, a residual connection is employed to merge the two embeddings, enabling the final local 3D embeddings to incorporate self- and cross-attention. 
Layer normalization is applied to the combined embeddings to ensure scale consistency by
\begin{equation}
\small
   \mathcal{\bar{F}}^{L} = \text{layerNorm}\left(\mathcal{\bar{F}}^{self} + \mathcal{\bar{F}}^{cross}\right) \in \mathbb{R}^{27 \times d_{out}}.
\end{equation}
Each local 3D embedding can be represented by $\bar{F}^{L}_{j} \in \mathbb{R}^{1 \times d_{out}}$, where $j$ denotes the local block index.

\subsection{Pseudo-labeling for 3D Reasoning}
\label{subsec:pseudo-label}
Global and local text labels are required for 3D reasoning. 
However, such labels are not provided in the benchmark datasets~\cite{ShapeNet,ModelNet}, and manually annotating large-scale datasets is costly. 
A pre-trained large language model (LLM), such as a generative pre-trained transformer (GPT)~\cite{GPT3}, can generate pseudo-text labels for 3D reasoning to address this problem.

\noindent \textbf{Object Pseudo-labeling (Global Branch).} 
The proposed pseudo-label generation comprises two prompt insertion stages: guidance and query.
First, guidance prompts are inserted, including detailed instructions, such as required response formats, necessary elements, formats to avoid, and examples of good answers. 
These guidance prompts are inserted only once and shared across all objects. 
No response from the LLM is required at this stage. 
The prompts play a critical role in enhancing the diversity and purposefulness of the generated responses. 
For example, the prompt instructs the LLM not to include the direct name or label of the object in its responses because we aim to perform the 3D reasoning task. 
Instead, the prompt guides the LLM to ask indirect questions that help infer the object, such as its purpose, function, structure, or usage.
Next, the query prompt ``Give a one-sentence description of \{\texttt{\small classname}\}," is used to obtain a response from the LLM, which is collected as the reasoning pseudo-label for the global 3D object.

\begin{figure}
  \begin{subfigure}[b]{0.32\columnwidth}
    \includegraphics[width=\linewidth]{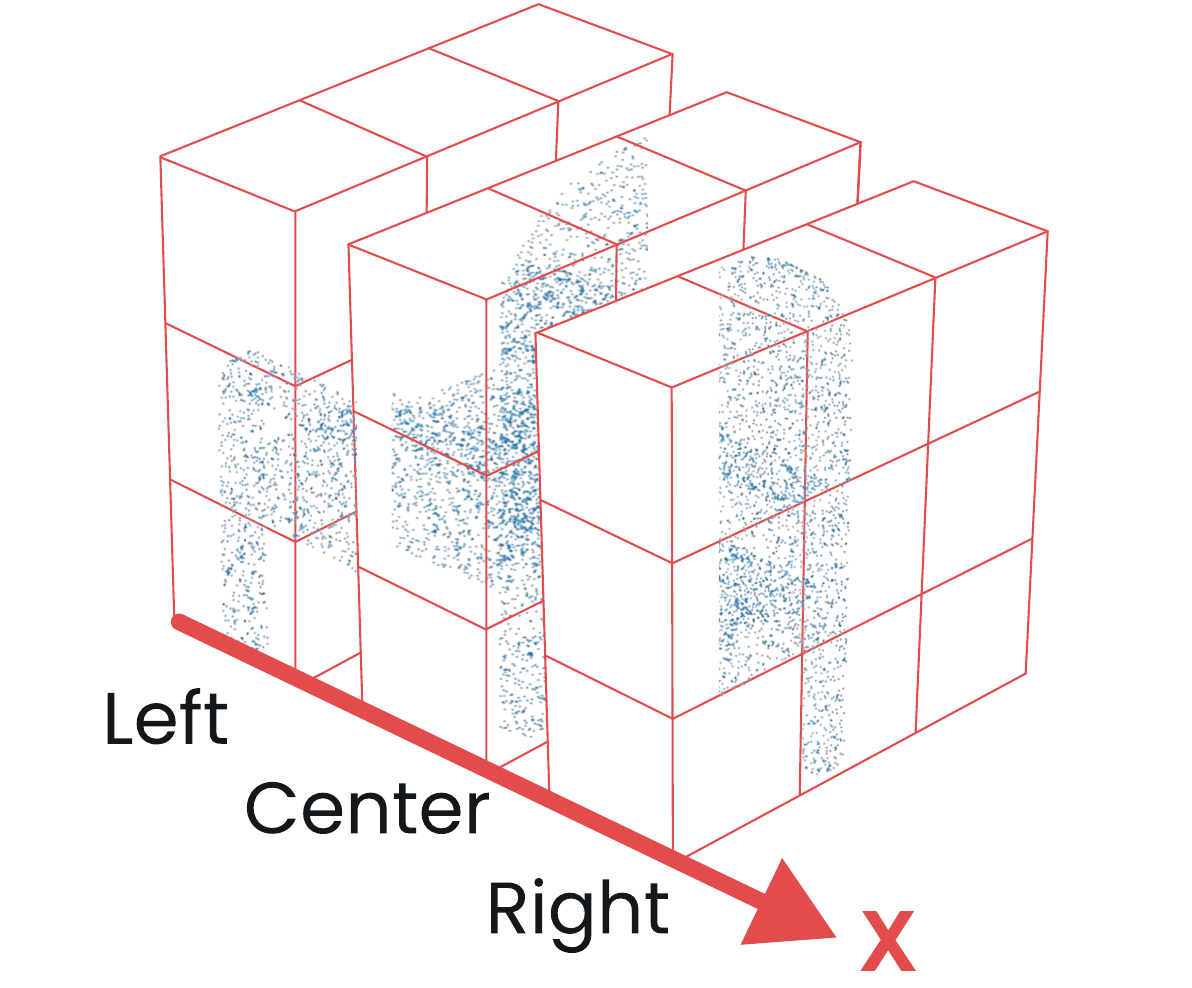}
    \caption{X-axis positions: \{left, center, right\}}
  \end{subfigure}
  \hfill 
  \begin{subfigure}[b]{0.32\columnwidth}
    \includegraphics[width=\linewidth]{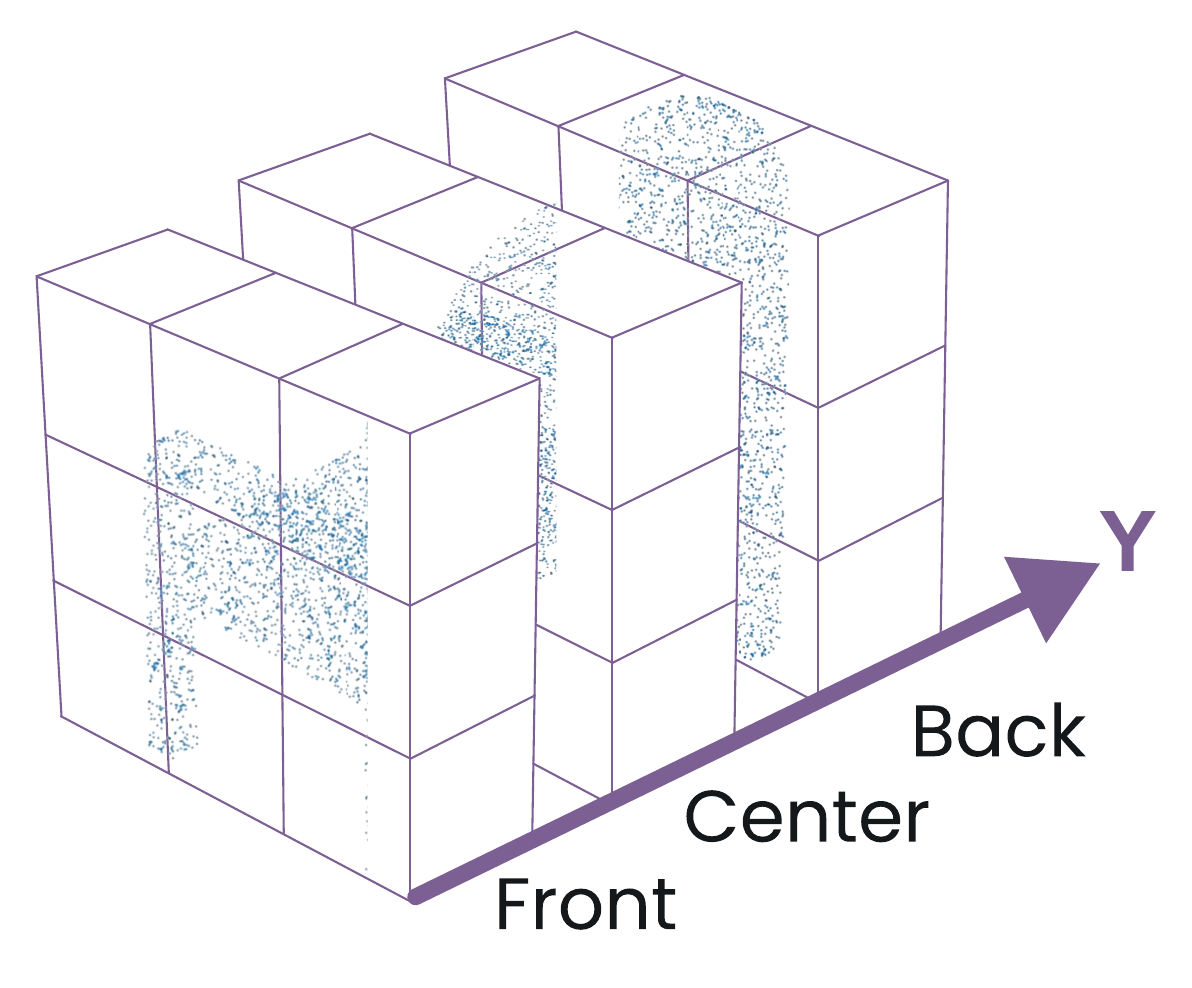}
    \caption{Y-axis positions: \{front, center, back\}}
  \end{subfigure}
  \hfill 
  \begin{subfigure}[b]{0.32\columnwidth}
    \includegraphics[width=\linewidth]{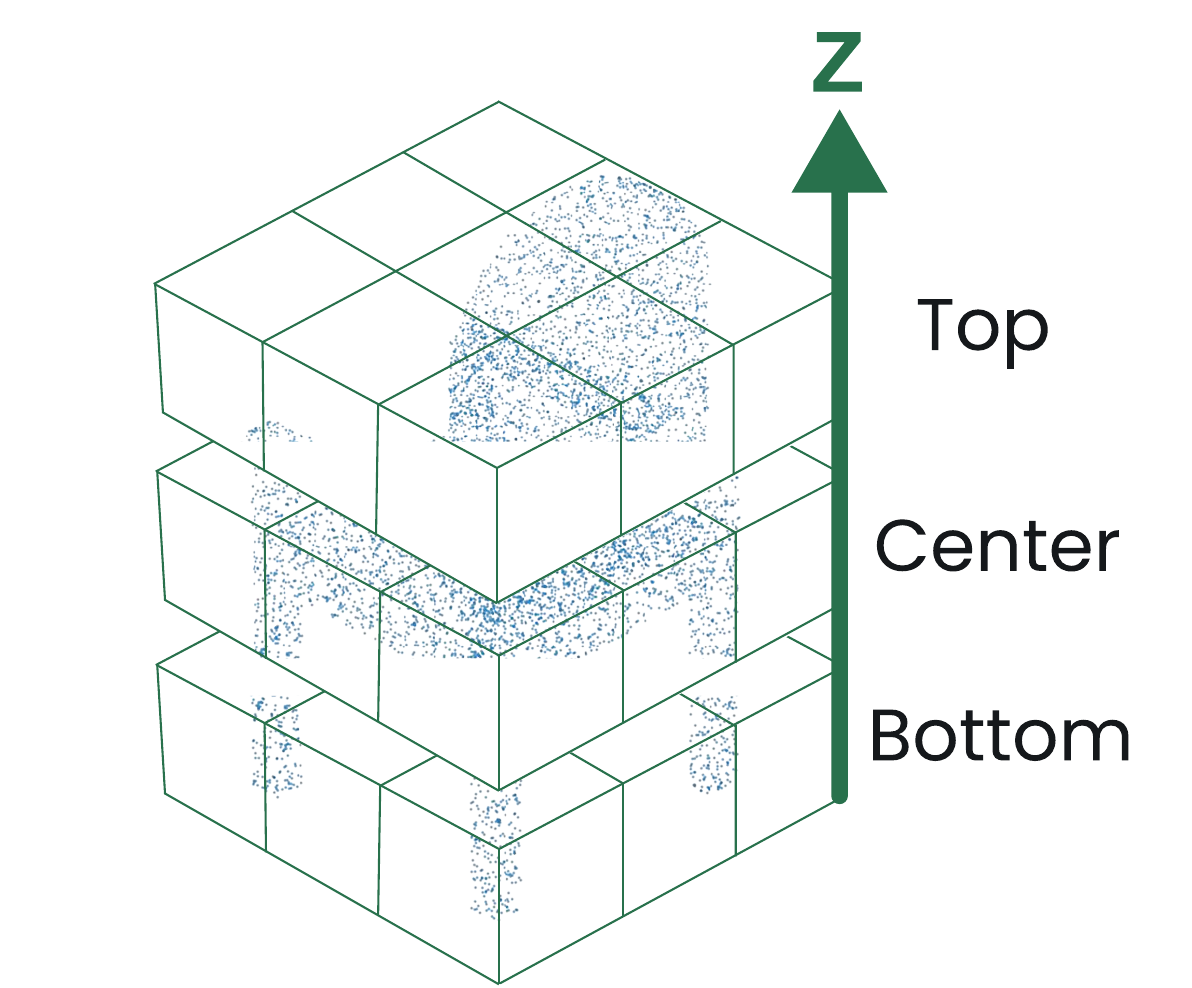}
    \caption{Z-axis positions: \{bottom, center, top\}}
  \end{subfigure}
  \caption{\textbf{Three intuitive positions \{\texttt{\small pos}\} for each axis}}
  \label{fig:pseudolabeling}
  \vspace{-10pt}
\end{figure}

\noindent \textbf{Local Block Pseudo-labeling (Local Branch).}
Reasoning-text labels for $3\times3\times3$ blocks are required to describe all local blocks.
However, extracting the labels for each of the 27 blocks via the LLM is ambiguous and inefficient.
To address this problem, we divide the 3D object into three intuitive positions~\{\texttt{\small pos}\} along each of the axes as described in Fig.~\ref{fig:pseudolabeling}.
Then, we create 9 query prompts using the following format: ``For the \{\texttt{\small pos}\} region of the \{\texttt{\small x,y,z}\}-axis of \{\texttt{\small classname}\}, provide EXACTLY ONE sentence describing its structural and functional features without referring to its position".
This approach is clearer and more effective than creating 27 individual query prompts. 
Despite using only 9 labels for the 27 local blocks, the novel local loss function proposed in Sec.~\ref{sec:loss} allows training 3D reasoning networks that understand all local blocks.

We define the global text label as $\text{Text}^{G}$ and the local text labels as $\text{Text}^{L}_k$, where $k=1,2,\dots,9$.
All prompts, including guidance and query prompts for global and local blocks, are provided in the supplementary materials.

\subsection{Text Embedding}
\label{subsec:text_embedding}
We further extract embedding features from the text labels through the pre-trained GPT text encoder ${E}_{T}$ by ${T}^{G}={E}_{T}(\text{Text}^{G}) \in \mathbb{R}^{d_{E_T}}$ and ${T}^{L}_k={E}_{T}(\text{Text}^{L}_k) \in \mathbb{R}^{d_{E_T}}$.
We employ the text-embedding-3-small model in GPT-3.5~\cite{GPT3} for the text encoder.
To align text embeddings with visual embeddings, we add a linear projection layer $\mathbf{W}^T\in \mathbb{R}^{d_{E_T} \times d_{out}}$ for global and local text embeddings by 
\begin{equation}
\begin{split}
    \bar{T}^{G} & = {T}^{G}\cdot \mathbf{W}^T \in \mathbb{R}^{1 \times d_{out}}, \\ 
    \bar{T}^{L}_{k} & = {T}^{L}_k\cdot \mathbf{W}^T \in \mathbb{R}^{1 \times d_{out}}.
\end{split}
\end{equation}
Finally, the global and local text embeddings~$(\bar{T}^{G},\bar{T}^{L}_{k})$ have the same dimensionality $d_{out}$ as the global and local 3D embeddings~$(\bar{F}^{G},\bar{F}^{L}_{j})$.

\subsection{Loss Functions}
\label{sec:loss}
This work designs global and local losses using an InfoNCE ~\cite{InfoNCELoss} loss to learn a shared embedding space for 3D embeddings~($\bar{F}$) and text embeddings~($\bar{T}$). 
The global loss between two global embeddings is designed as follows:
\begin{equation}
\small
    \mathcal{L}_{global} = -\frac{1}{N}\sum^N_{l=1}\left[\log\frac{f\left(\bar{F}_{(l)}^{G},\bar{T}_{(l)}^{G}\right)/\tau}{\sum^N_{m=1}f\left(\bar{F}_{(l)}^{G},\bar{T}_{(m)}^{G}\right)/\tau}\right],
\label{eq:global_loss}
\end{equation}
where $N$ represents the batch size, $l$ and $m$ denote the object indices in the batch, and $\tau$ indicates a scaling factor.
The function $f(\mathbf{a},\mathbf{b})=\exp(\tfrac{\mathbf{a}}{\mathbf{\|a\|}}\cdot\tfrac{\mathbf{b}}{\mathbf{\|b\|}})$ denotes a similarity measure between two embedding vectors.
This global loss encourages positive pairs $(\bar{F}_{(l)}^{G},\bar{T}_{(l)}^{G})$ to have higher similarity while pushing apart negative pairs $(\bar{F}_{(l)}^{G},\bar{T}_{(m)}^{G})$ for $l\neq m$.

The local loss between local embeddings is designed as follows:
\begin{equation}
\small
    \mathcal{L}_{local} = -\mathbb{E}\left[\log\frac{\sum^{27}_{j=1}\sum^{9}_{k=1}{P(j,k)f(\bar{F}^{L}_j,\bar{T}^{L}_k)}/\tau}{\sum^{27}_{j=1}\sum^{9}_{k=1}{f(\bar{F}^{L}_j,\bar{T}^{L}_k)}/\tau}\right], 
\label{eq:local_loss}
\end{equation}
where $j$ represents the index of a local block, and $k$ denotes the index of a local text label.
The function $P(j,k)$ is a pairwise similarity indicator that determines whether two embeddings $\bar{F}^{L}_j$ and $\bar{T}^{L}_k$ form a positive or negative pair. 
When the $j$th block contains no point cloud or too few points, it is excluded from the local loss function.
Unlike global loss, which considers pairs between different objects, local loss focuses on positive and negative pairs in the same object.

\begin{figure}[t]
   \centering 
   \includegraphics[width=0.95\linewidth]{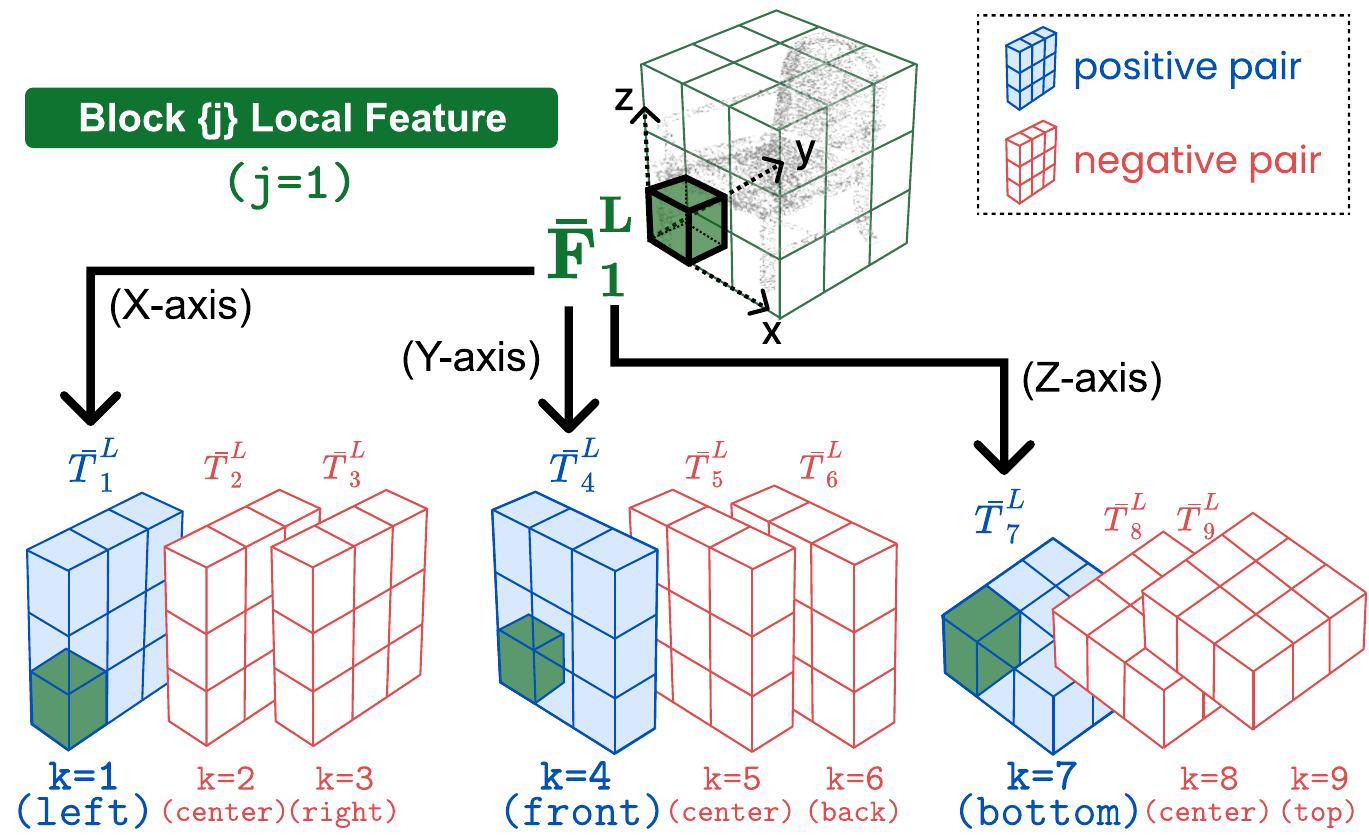}
   \caption{\textbf{Examples of positive and negative pairs assignments using $P(j,k)$.} Each local 3D feature corresponds to three positive and six negative text-embedding pairs.}
   \label{fig:matching_pair}
\end{figure}

The local blocks of the 3D object consist of 27 blocks, whereas only 9 text labels were generated, as described in Sec.~\ref{subsec:pseudo-label}.
To assign positive and negative relationships between two local embeddings $\bar{F}^{L}_j$ and $\bar{T}^{L}_k$,
we effectively match them as illustrated in Fig.~\ref{fig:matching_pair}.
For example, if a local 3D embedding has a block index of $j=1$, this index corresponds to the location ($x=1$, $y=1$, $z=1$) in the $3\times3\times3$ block structure. It can be associated with the first text embeddings along each axis. Then, the text embeddings $\bar{T}^L_k$ for $k=1,4,7$ form three positive pairs with the local 3D embedding $\bar{F}^L_j$, where $j=1$. Other pairs are negatives.
Thus, the indicator function is defined as follows:
\begin{equation}
\small
	P(j,k)=\begin{cases} 1\quad \text{if}~ (j,k)~\text{is a positive pair} \\ 0\quad \text{otherwise} \end{cases}.
\end{equation}
This indicator is binary, assigning 1 to positive pairs and 0 to negative pairs  $P(j,k)\in \{1,0\}$.
Through the indicator, the local loss in Eq.~\ref{eq:local_loss} trains the local embedding space by encouraging positive pairs to have higher similarity than negative pairs, called the hard local loss.

\begin{figure}[t]
   \centering 
   \includegraphics[width=1\linewidth]{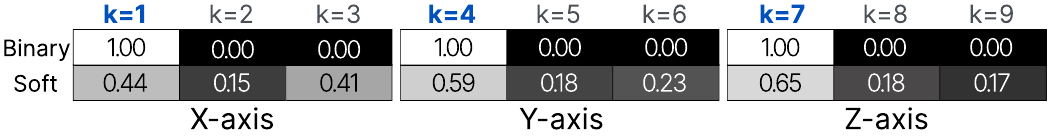}
   \caption{\textbf{Comparison of the binary and soft indicators,} where $j=1$ and the \{\texttt{\small classname}\} is ``chair."}
   \label{fig:hard_vs_soft}
\end{figure}

Although the local loss with the binary indicator can train the network, it does not account for symmetrical structures in certain objects. 
For instance, the left and right armrests are symmetrically positioned in a chair along the $x$ -axis, sharing similar 3D features and text labels. 
However, despite their high similarity, the binary indicator categorizes them as a negative pair.
To address this problem, we propose a similarity-based soft indicator by computing similarities between text embeddings along each axis. 
First, the positive indices $k$ with each $j$ are calculated for each axis based on $P(j,k)$.
For each axis, the similarities of the text embeddings are represented by
\begin{equation}
\small
\begin{split}
    S_j(T^L_{pos-x}, T^L_{k}),\quad \text{where} \quad k=\{1,2,3\}, \\
    S_j(T^L_{pos-y}, T^L_{k}),\quad \text{where} \quad k=\{4,5,6\}, \\
    S_j(T^L_{pos-z}, T^L_{k}),\quad \text{where} \quad k=\{7,8,9\},
\end{split}
\end{equation}
where ${pos-{x}}$ represents the index of the positive pair $k$ corresponding to the $j$th local 3D embedding along the $x$-axis.
The similarity function $S_j$ is calculated by the softmax function as follows:
\begin{equation}
\small
S_j(T^L_{pos-a}, T^L_{k}) = \frac{f(T^L_{pos-a},T^L_k)}{\sum_k f(T^L_{pos-a},T^L_k)},
\end{equation}
where $f(\cdot)$ denotes the same similarity function as in Eq.~\ref{eq:global_loss} and $a \in \{x,y,z\}$.
Unlike $P(j,k)$, which only assigns binary values, this indicator provides smooth weighting based on the similarity of text embeddings, as illustrated in Fig.~\ref{fig:hard_vs_soft}.
The soft local loss is defined by replacing the binary indicator $P(j,k)$ in the local loss function (Eq.~\ref{eq:local_loss}) with the soft indicator $S_j(T^L_{pos-a}, T^L_{k})$. 

Finally, the total loss for PointCubeNet is defined as the sum of the global and local losses:
\begin{equation}
   \mathcal{L}_{total} = \mathcal{L}_{global} + \mathcal{L}_{local}.
\end{equation}
The two losses are combined in a one-to-one ratio, and we tested the hard and soft local losses as part of the total loss in the experiments.
All parameters in PointCubeNet are learned during training, except for the text embedding networks.

\section{Experimental Results}
\subsection{Datasets and Settings}
\noindent \textbf{Datasets.} The 3D point cloud datasets \texttt{ModelNet}~\cite{ModelNet} and \texttt{ShapeNet}~\cite{ShapeNet} were  used in the experiments.
\texttt{ModelNet}~\cite{ModelNet} consists of about 12,300 objects across 40 categories.  
Each object comprises 2048 sampled points.
Most objects in \texttt{ModelNet} were not pre-aligned, so we manually adjusted them to ensure consistent orientation.
\texttt{ShapeNet}~\cite{ShapeNet} consists of about 51,300 objects across 55 categories, which is a larger dataset than \texttt{ModelNet}. 
Each object comprises 8,192 densely sampled points.
For evaluation, 40 categories were used for 3D classification and reasoning, while 15 were used for zero-shot part-level 3D reasoning.
We used all sampled points (8192 for \texttt{ShapeNet} and 2048 for \texttt{ModelNet}) in those datasets for experiments.

\noindent \textbf{Settings.}
We conducted two quantitative evaluations: 3D classification and 3D reasoning.
In 3D classification, the global text label corresponds to the object category ``\{\texttt{\small classname}\}'', and the local text labels are defined as the ``\{\texttt{\small pos}\} region of the \{\texttt{\small x,y,z}\}-axis for \{\texttt{\small classname}\}.'' 
For 3D reasoning, several similar pseudo-labels per class were generated based on the proposed pseudo-labeling in Sec.~\ref{subsec:pseudo-label} for global and local labels. 
One was randomly selected among the labels to enhance the diversity of the training labels. 
Similarities between the global 3D features and global text embeddings were calculated, and the average top-1 accuracy was measured to evaluate the 3D classification and reasoning. 
Furthermore, the 3D part-level reasoning was qualitatively evaluated. 
Training labels were paraphrased to generate user text prompts for the model test to prevent the trained model from simply memorizing specific reasoning labels.  
The user text prompt and a 3D point cloud were input into the trained networks, and the similarity between the query and 27 local 3D features was measured to visualize the heat-map, as depicted in Fig.~\ref{fig:part-level_reasoning}.

\begin{table}[t]
   \footnotesize
   \centering
   \renewcommand{\arraystretch}{0.8} 
   \setlength\tabcolsep{2.2pt}
   \begin{tabular}{l|c|c|c}
   \Xhline{3\arrayrulewidth}
   Network structure  & Loss function          & 3D classification & 3D reasoning       \\ \hline
   only global branch & global                 & 83.59\%           & 80.48\%            \\ \hline
   w/o Attns          & global+local (hard)    & 82.19\%           & 80.73\%            \\ \hline
   only self-Attn     & global+local (hard)    & 83.62\%           & 81.11\%            \\ \hline
   only cross-Attn    & global+local (hard)    & 84.02\%           & 79.40\%            \\ \hline
   all (hard)         & global+local (hard)    & \tb{84.52\%}      & \tb{81.30\%}       \\ \hline
   all (soft)         & global+local (soft)    & \tR{85.06\%}      & \tR{81.43\%}       \\ 
   \Xhline{3\arrayrulewidth}
   \end{tabular}
   \caption{\textbf{Effects of the proposed methods.} The best and second-best scores are marked in \tR{red} and \tb{blue}.}
   \label{tab:exp_proposed}
   \end{table}

\begin{figure}[t]
  \begin{subfigure}[b]{0.45\columnwidth}
    \includegraphics[width=\linewidth]{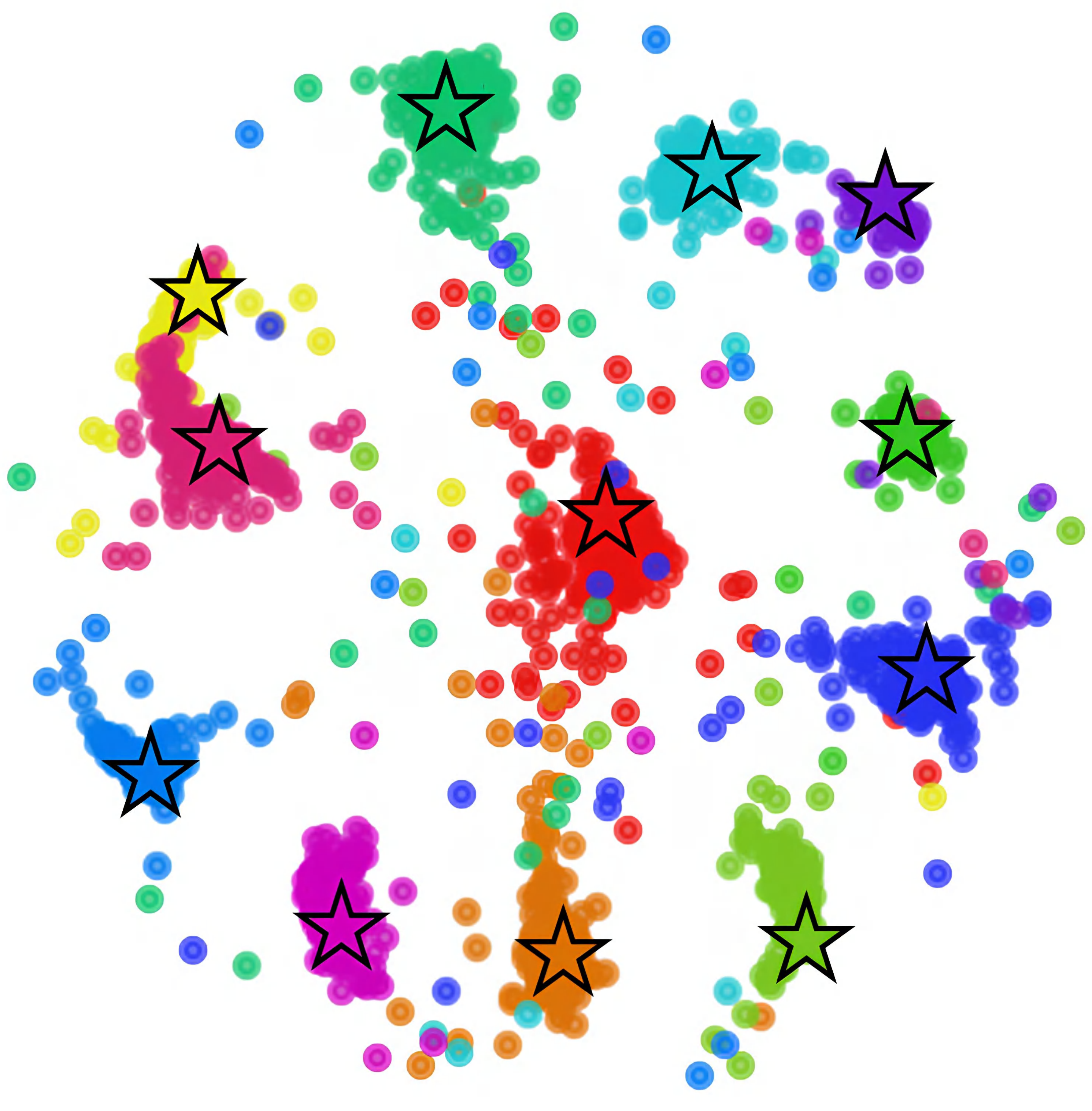}
    \caption{Global branch only}
  \end{subfigure}
  \hfill 
  \begin{subfigure}[b]{0.44\columnwidth}
    \includegraphics[width=\linewidth]{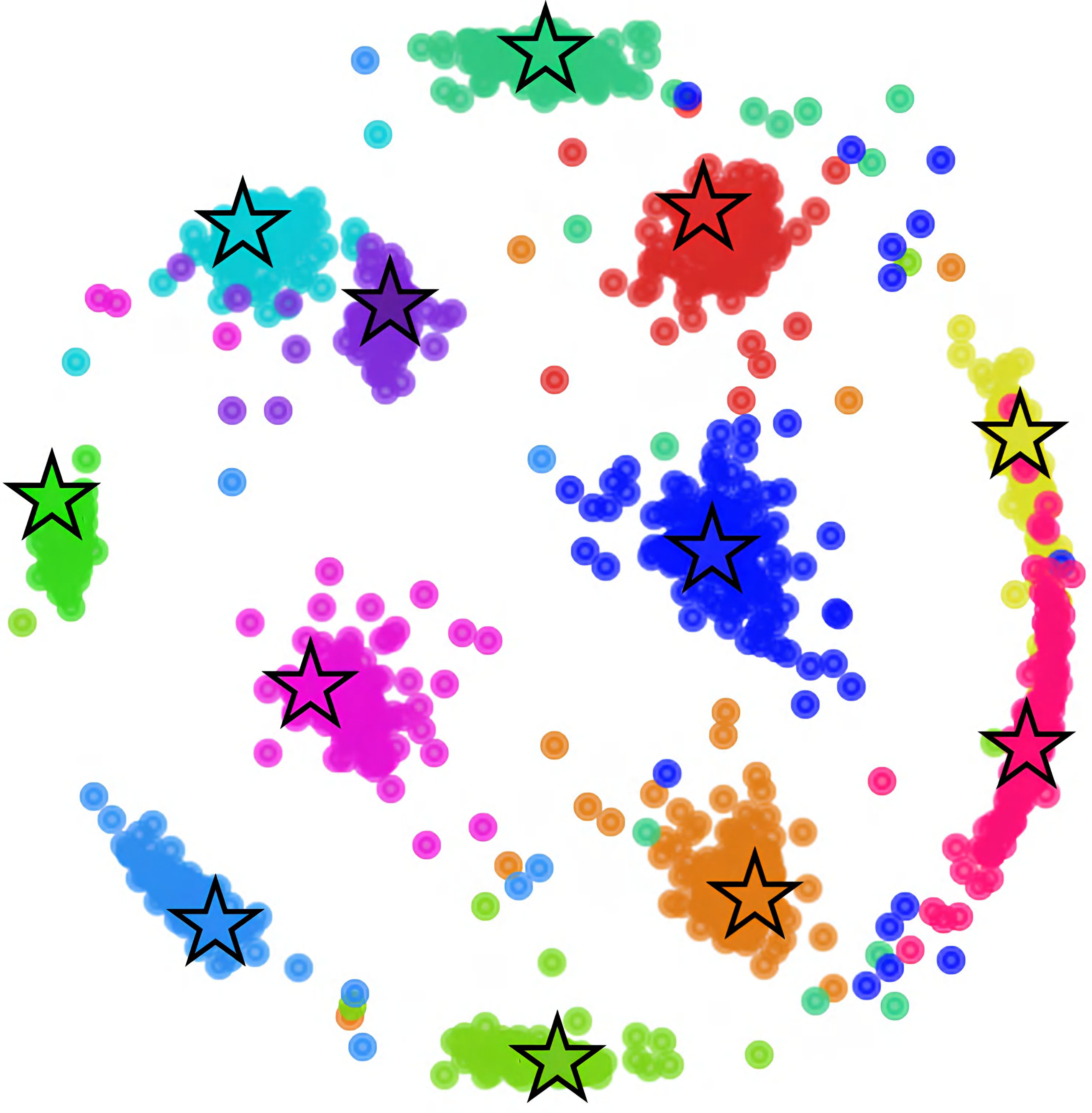}
    \caption{Global and local branches}
  \end{subfigure}
  \caption{\textbf{Visualization of global embeddings.} Colors represent different classes. Dots indicate 3D feature embeddings; stars represent text embeddings.}
  \vspace{-5pt}
  \label{fig:visual_global_embbed}
\end{figure}

\subsection{Effectiveness of the Proposed Methods}
We evaluated network structures and loss functions, including those for 3D classification and reasoning tasks, to validate the effectiveness of the proposed methods. 
We defined different versions of the proposed methods as follows:
$\bullet$ only global branch: A global branch without a local branch.
$\bullet$ w/o Attns: Without self- and cross-attentions in the local 3D feature extractor.
$\bullet$ only self-Attn or cross-Attn: With only self- or cross-attention.
$\bullet$ all (hard) or all (soft): With both attention mechanisms in the local 3D feature extractor, optimized using the hard or soft local loss.
We utilized DGCNN~\cite{DGCNN} for the 3D encoder in PointCubeNet and evaluated the methods on the \texttt{ShapeNet}~\cite{ShapeNet} dataset.

Table~\ref{tab:exp_proposed} reveals that the 3D classification and reasoning performance gradually improves as the local branch and its soft loss are incorporated.
Although 3D classification and reasoning can be performed using only the global branch during testing, training the local branch further improves performance over training with only the global branch.
These results demonstrate that considering local parts of the object enhances the understanding of whole 3D objects. 
Moreover, the proposed soft loss further improves the performance of PointCubeNet by enabling more flexible training of the local branch and enhancing the handling of symmetric objects. 
Figure~\ref{fig:visual_global_embbed} visualizes the global 3D features and text embeddings for training with only the global branch versus the full PointCubeNet.
When all branches are trained together, the distribution of global embeddings becomes more discriminative, highlighting the importance of local information in 3D object analysis.

\subsection{Performance Comparison}

We compared 3D classification and reasoning in the proposed method with those of state-of-the-art methods. 
For these evaluations, we utilized DGCNN~\cite{DGCNN} as the baseline 3D encoder for PointCubeNet.
Table \ref{tab:classification} summarizes the classification performances.
In both datasets, PointCubeNet improved the classification performance of its baseline.
ULIP\cite{ULIP}, a CLIP-based method, achieved the best performance on the \texttt{ModelNet} dataset.
However, our method also demonstrated competitive performance on \texttt{ModelNet} (93.6\%) and \texttt{ShapeNet} (85.06\%).
Unlike all other compared methods, PointCubeNet does not train classification heads yet achieved superior performance in classification.

\begin{table}[t]
   \footnotesize
   \centering
   \renewcommand{\arraystretch}{1} 
   \begin{tabular}{r|c|c}
   \Xhline{3\arrayrulewidth}
   Methods $\symbol{92}$Dataset & \texttt{ModelNet}~\cite{ModelNet}    & \texttt{ShapeNet}~\cite{ShapeNet}      \\ \hline
   VoxNet~\cite{maturana2015voxnet}                 & 83.0\%  & --  \\ \hline
   PointNet~\cite{PointNet}               &  89.2\% & 81.21\%  \\ \hline
   PointNet++~\cite{PointNet2}             & 91.9\%   &  83.02\% \\ \hline
   PointCNN~\cite{li2018pointcnn}              & 92.2\%  &  83.24\% \\ \hline
   PCNN~\cite{atzmon2018point}                   & 92.3\%  & 83.52\%  \\ \hline
   DGCNN~\cite{DGCNN}                   & 92.9\%  & 83.59\%  \\ \hline
   $\dagger$PointMLP~\cite{marethinking}               & \tb{94.5\%}   & 82.64\%  \\ \hline
   $\dagger$ULIP~\cite{ULIP}                   & \tR{94.7\%}  & \tR{85.21\%}  \\ \hline \hline
   Ours -- all (hard)                & 93.0\%  &  84.52\%  \\ \hline
   Ours -- all (soft)                 & 93.6\%  &  \tb{85.06\%} \\ 
   \Xhline{3\arrayrulewidth}
   \end{tabular}        
   \caption{\textbf{Comparison of 3D classification performance.} \\ $\dagger$ denotes 2D projection-based methods using CLIP~\cite{CLIP}. The best and second-best are marked in \tR{red} and \tb{blue}.}
   \label{tab:classification}
\end{table}

\begin{table}[t]
   \footnotesize
   \centering
   \renewcommand{\arraystretch}{0.8} 
   \begin{tabular}{r|c|c}
   \Xhline{3\arrayrulewidth}
   Methods $\symbol{92}$Dataset & \texttt{ModelNet}~\cite{ModelNet}     & \texttt{ShapeNet}~\cite{ShapeNet}      \\ \hline
   $\dagger$PointCLIPv2~\cite{PointCLIPV2}            & 63.86\%  &  62.77\% \\ \hline
   $\dagger$ULIP~\cite{ULIP}                   & --  & 68.83\%  \\ \hline 
   PointNet~\cite{PointNet} + \textit{GB} & 84.29\%  & 79.65\%  \\ \hline
   PointNet++~\cite{PointNet2} + \textit{GB} &  83.89\%  & 76.85\%  \\ \hline
   DGCNN~\cite{DGCNN} + \textit{GB} & 85.71\%  & 80.48\%  \\ \hline \hline
   Ours -- all (hard)      & \tb{86.28\%}  & \tb{81.30\%}  \\ \hline
   Ours -- all (soft)      & \tR{88.07\%} & \tR{81.43\%}  \\ 
   \Xhline{3\arrayrulewidth}
   \end{tabular}
   \caption{\textbf{Comparison of 3D reasoning performance.} $\dagger$ denotes 2D projection-based methods using CLIP~\cite{CLIP}. \textit{GB} means applying the global branch for the 3D reasoning task.}
   \label{tab:reasoning}
   \end{table}

For 3D reasoning, point cloud-based methods, such as the PointNet series~\cite{PointNet, PointNet2} and DGCNN~\cite{DGCNN}, were initially incapable of reasoning due to their network structures.
To address this problem, we modified these methods by integrating the global branch~(\textit{GB}), enabling them to perform 3D reasoning.
For CLIP-based methods (ULIP~\cite{ULIP} and PointCLIPv2~\cite{PointCLIPV2}), we froze their text embedding modules and fine-tuned the stacked MLP layers, similar to PointCubeNet.
However, ULIP~\cite{ULIP} relies heavily on rendered images for training, but only \texttt{ShapeNet} renders were available online, making reasoning experiments infeasible.
For 3D reasoning, the proposed PointCubeNet outperforms all other methods on both datasets, as shown in Tab.~\ref{tab:reasoning}.
Notably, CLIP-based methods\cite{CLIP, PointCLIPV2} performed worse in reasoning than classification, suggesting that the text encoder in CLIP~\cite{CLIP} was not trained well for reasoning.

\begin{table}[t]
\footnotesize
\centering
\renewcommand{\arraystretch}{0.8} 
\begin{tabular}{r|cc|cc}
\Xhline{3\arrayrulewidth}
Tasks                         & \multicolumn{2}{c|}{3D classification} & \multicolumn{2}{c}{3D reasoning} \\ \hline
Methods$\symbol{92}$Dataset & \multicolumn{1}{c|}{\texttt{S$\rightarrow$M}} & \texttt{M$\rightarrow$S}      & \multicolumn{1}{c|}{\texttt{S$\rightarrow$M}}         &  \texttt{M$\rightarrow$S}   \\ \hline
PointNet++~\cite{PointNet2} + \textit{GB}   & \multicolumn{1}{c|}{83.17\%}& 53.20\% & \multicolumn{1}{c|}{78.75\%} & 66.88\%\\ \hline
DGCNN~\cite{DGCNN} + \textit{GB}           & \multicolumn{1}{c|}{87.22\%}& 63.83\%  & \multicolumn{1}{c|}{89.31\%} & 73.19\%\\ \hline 
PointNet~\cite{PointNet} + \textit{GB}   & \multicolumn{1}{c|}{92.14\%}&  87.52\%  & \multicolumn{1}{c|}{\tb{91.65\%}} & 90.13\% \\ \hline \hline
Ours -- all (hard)               & \multicolumn{1}{c|}{\tb{93.00\%}} & \tb{87.94\%} & \multicolumn{1}{c|}{91.25\%}  &  \tb{90.47\%}   \\ \hline
Ours -- all (soft)        & \multicolumn{1}{c|}{\tR{93.37\%}}& \tR{88.62\%} & \multicolumn{1}{c|}{\tR{91.77\%}}  & \tR{90.73\%}  \\ 
      \Xhline{3\arrayrulewidth}
\end{tabular}
\caption{\textbf{Cross-domain (source$\rightarrow$target) evaluation.} \texttt{S}:\texttt{\small ShpaeNet}, \texttt{M}:\texttt{\small ModelNet}. \textit{GB} means applying the global branch for the 3D reasoning task.}
\label{tab:cross_domain}
\end{table}

Additionally, we conducted cross-domain experiments for both tasks, as presented in Tab.~\ref{tab:cross_domain}.
The \texttt{ShapeNet} and \texttt{ModelNet} datasets share 13 common classes, so we performed cross-domain testing only on these overlapping classes.
Among the tested 3D point cloud-based methods, PointNet~\cite{PointNet} achieved the highest performance and was chosen as the baseline for PointCubeNet.
PointCubeNet consistently outperformed the baseline across all tasks in cross-domain experiments. 
Notably, the proposed method maintained robust performance despite the significant size difference between the two datasets (e.g., 8192 points for ShapeNet vs. 2048 points for ModelNet).  
This result demonstrates the robustness and stability of the proposed method across diverse point cloud densities.

\begin{figure*}[t]
   \centering
   \includegraphics[width=1\linewidth]{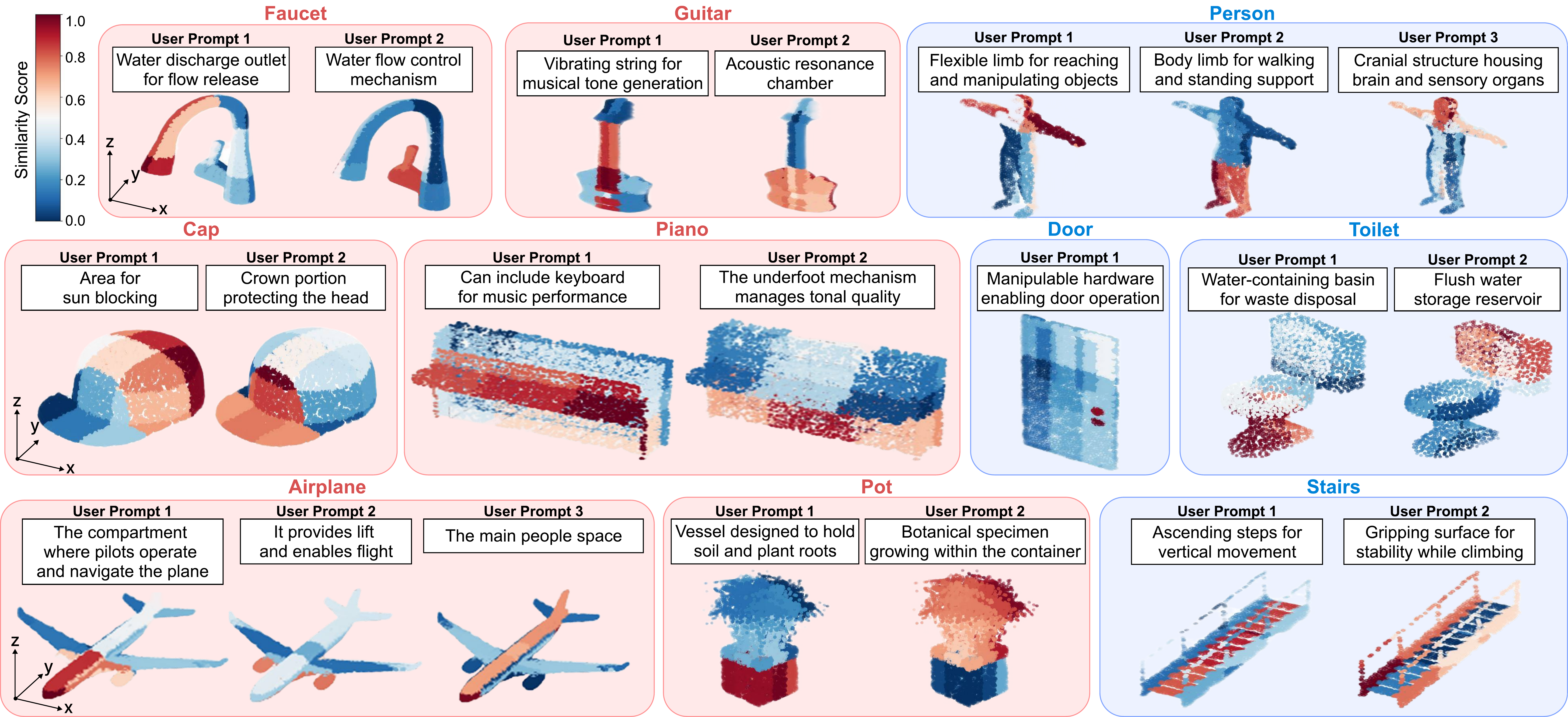}
   \caption{\textbf{Results for 3D part-level reasoning on \tr{\texttt{ShapeNet}}~\cite{ShapeNet} and \tb{\texttt{ModelNet}}~\cite{ModelNet}.} User prompts are newly generated or paraphrased from the training labels. In 3D part-level reasoning, the similarities between the user prompt embeddings and 3×3×3 3D feature embeddings were computed. Best viewed in color and additional results in the supplementary materials.}  
   \vspace{-5pt}
   \label{fig:part-level_reasoning}
\end{figure*}

\begin{figure*}[t]
   \centering
   \includegraphics[width=1
   \linewidth]{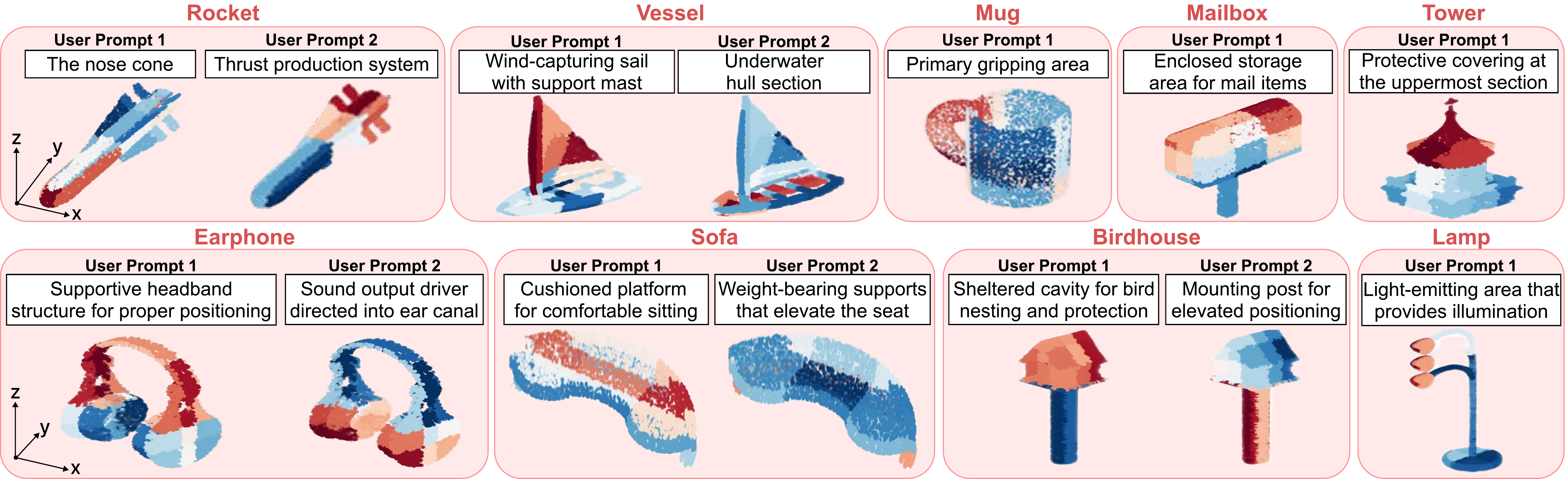}
   \caption{\textbf{Results for zero-shot 3D part-level reasoning on \texttt{ShapeNet}\cite{ShapeNet}.} Although tested categories are unseen, PointCubeNet correctly performs 3D part-level reasoning. Best viewed in color.}
   \label{fig:zero-shot_part-level_reasoning}
\end{figure*}

\subsection{3D Part-level Reasoning}
The local branch structure of the proposed PointCubeNet allows 3D part-level reasoning. 
A recent study~\cite{PARIS3D} explored 3D part reasoning segmentation using supervised learning with reasoning annotations for 3D object parts. 
In contrast, PointCubeNet generates the annotations via proposed pseudo-labeling, as described in Sec.~\ref{subsec:pseudo-label}, enabling unsupervised part-level reasoning.
As illustrated in Fig.~\ref{fig:part-level_reasoning}, PointCubeNet can compute similarities between a user text prompt and 27 local blocks.
We re-generated and paraphrased the reasoning user prompts from the training text labels for this evaluation. 
Even when the user prompts do not explicitly include the class name but provide indirect descriptions, this approach successfully 
performs part-level reasoning. 27 local blocks are sufficient to represent the detailed parts of an object.

\begin{figure}[t]
   \centering 
   \includegraphics[width=1\linewidth]{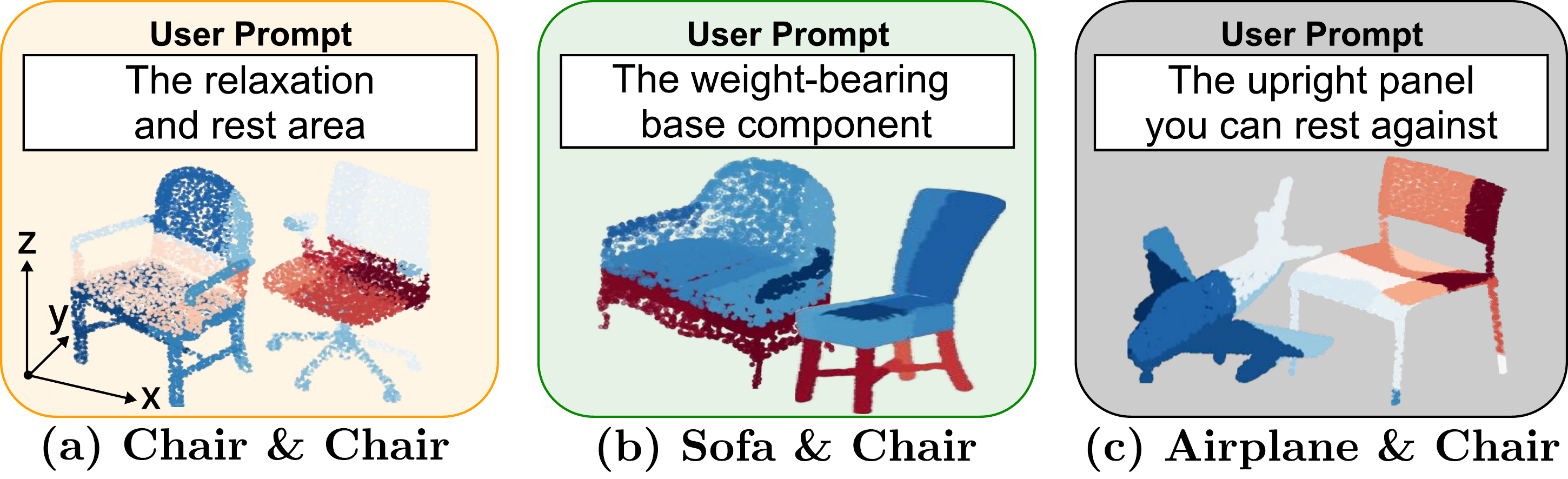}
   \caption{\textbf{Part-level reasoning with multi-object inputs.} Two objects belonging to the same, similar, or different classes are input into 3x3x3 blocks. The results of PointCubeNet are reasonable across all scenarios.}
   \label{fig:multi_object}
\end{figure}

Moreover, PointCubeNet also enables zero-shot part-level reasoning.
As displayed in Fig.~\ref{fig:zero-shot_part-level_reasoning}, the tested 3D objects belong to unseen categories during the network training stage.
Similar categories, such as chairs and beds, exist in the training data for the sofa class. However, despite the absence of visually or functionally similar classes in the training set, PointCubeNet successfully performed part reasoning for other objects.
These results demonstrate the zero-shot generalizability and robustness of the proposed method to variation.
As depicted in Fig.~\ref{fig:multi_object}, PointCubeNet correctly inferred part reasoning even when multiple objects were input into a single cube.
This finding indicates that PointCubeNet does not simply learn the absolute positions of local parts but accurately captures the relationships between local parts and the given text prompt.
More qualitative results are provided in the supplementary materials.

\section{Conclusions and Future Works}
In this paper, we proposed PointCubeNet, a multi-modal 3D understanding framework capable of performing 3D classification, reasoning, and part-level reasoning.
To achieve this aim, we designed a local branch to capture the local features of 3D objects effectively. 
We introduced a pseudo-labeling method and a novel local loss function to enhance learning further. 
The experimental results validated the effectiveness of PointCubeNet. 
Notably, understanding 3D object parts improved the comprehension of entire 3D objects. 
Furthermore, the local branch enabled unsupervised 3D part-level reasoning without human annotations, producing reliable and meaningful results.

However, the reliance of PointCubeNet on 3x3x3 local structures may lead to prediction discontinuities at local block boundaries. 
Additionally, this approach struggles to perform reasoning-based segmentation at the point cloud level. 
Future work can address these limitations of PointCubeNet via post-processing or developing an additional segmentation network for improved reasoning accuracy in an unsupervised manner.

{\small
\bibliographystyle{ieee_fullname}
\bibliography{egbib}

@inproceedings{xue2024ulip,
  title={Ulip-2: Towards scalable multimodal pre-training for 3d understanding},
  author={Xue, Le and Yu, Ning and Zhang, Shu and Panagopoulou, Artemis and Li, Junnan and Mart{\'\i}n-Mart{\'\i}n, Roberto and Wu, Jiajun and Xiong, Caiming and Xu, Ran and Niebles, Juan Carlos and others},
  booktitle={Proceedings of the IEEE/CVF Conference on Computer Vision and Pattern Recognition},
  year={2024}
}

@article{schuhmann2022laion,
  title={Laion-5b: An open large-scale dataset for training next generation image-text models},
  author={Schuhmann, Christoph and Beaumont, Romain and Vencu, Richard and Gordon, Cade and Wightman, Ross and Cherti, Mehdi and Coombes, Theo and Katta, Aarush and Mullis, Clayton and Wortsman, Mitchell and others},
  journal={Advances in neural information processing systems},
  volume={35},
  year={2022}
}

@inproceedings{li2022blip,
  title={Blip: Bootstrapping language-image pre-training for unified vision-language understanding and generation},
  author={Li, Junnan and Li, Dongxu and Xiong, Caiming and Hoi, Steven},
  booktitle={International conference on machine learning},
  year={2022},
  organization={PMLR}
}

@inproceedings{jia2021scaling,
  title={Scaling up visual and vision-language representation learning with noisy text supervision},
  author={Jia, Chao and Yang, Yinfei and Xia, Ye and Chen, Yi-Ting and Parekh, Zarana and Pham, Hieu and Le, Quoc and Sung, Yun-Hsuan and Li, Zhen and Duerig, Tom},
  booktitle={International conference on machine learning},
  year={2021},
  organization={PMLR}
}

@inproceedings{szegedy2015going,
  title={Going deeper with convolutions},
  author={Szegedy, Christian and Liu, Wei and Jia, Yangqing and Sermanet, Pierre and Reed, Scott and Anguelov, Dragomir and Erhan, Dumitru and Vanhoucke, Vincent and Rabinovich, Andrew},
  booktitle={Proceedings of the IEEE conference on computer vision and pattern recognition},
  year={2015}
}

@inproceedings{he2016deep,
  title={Deep residual learning for image recognition},
  author={He, Kaiming and Zhang, Xiangyu and Ren, Shaoqing and Sun, Jian},
  booktitle={Proceedings of the IEEE conference on computer vision and pattern recognition},
  year={2016}
}

@inproceedings{su2015multi,
  title={Multi-view convolutional neural networks for 3d shape recognition},
  author={Su, Hang and Maji, Subhransu and Kalogerakis, Evangelos and Learned-Miller, Erik},
  booktitle={Proceedings of the IEEE international conference on computer vision},
  year={2015}
}

@inproceedings{qi2016volumetric,
  title={Volumetric and multi-view cnns for object classification on 3d data},
  author={Qi, Charles R and Su, Hao and Nie{\ss}ner, Matthias and Dai, Angela and Yan, Mengyuan and Guibas, Leonidas J},
  booktitle={Proceedings of the IEEE conference on computer vision and pattern recognition},
  year={2016}
}

@inproceedings{CLIP,
  title={Learning transferable visual models from natural language supervision},
  author={Radford, Alec and Kim, Jong Wook and Hallacy, Chris and Ramesh, Aditya and Goh, Gabriel and Agarwal, Sandhini and Sastry, Girish and Askell, Amanda and Mishkin, Pamela and Clark, Jack and others},
  booktitle={International conference on machine learning},
  year={2021},
  organization={PMLR}
}

@inproceedings{LISA,
  title={Lisa: Reasoning segmentation via large language model},
  author={Lai, Xin and Tian, Zhuotao and Chen, Yukang and Li, Yanwei and Yuan, Yuhui and Liu, Shu and Jia, Jiaya},
  booktitle={Proceedings of the IEEE/CVF Conference on Computer Vision and Pattern Recognition},
  year={2024}
}

@inproceedings{PARIS3D,
  title={PARIS3D: Reasoning-Based 3D Part Segmentation Using Large Multimodal Model},
  author={Kareem, Amrin and Lahoud, Jean and Cholakkal, Hisham},
  booktitle={European Conference on Computer Vision},
  year={2025},
  organization={Springer}
}

@inproceedings{PointBERT,
  title={Point-bert: Pre-training 3d point cloud transformers with masked point modeling},
  author={Yu, Xumin and Tang, Lulu and Rao, Yongming and Huang, Tiejun and Zhou, Jie and Lu, Jiwen},
  booktitle={Proceedings of the IEEE/CVF conference on computer vision and pattern recognition},
  year={2022}
}

@inproceedings{PointNet,
  title={Pointnet: Deep learning on point sets for 3d classification and segmentation},
  author={Qi, Charles R and Su, Hao and Mo, Kaichun and Guibas, Leonidas J},
  booktitle={Proceedings of the IEEE conference on computer vision and pattern recognition},
  year={2017}
}

@article{PointNet2,
  title={Pointnet++: Deep hierarchical feature learning on point sets in a metric space},
  author={Qi, Charles Ruizhongtai and Yi, Li and Su, Hao and Guibas, Leonidas J},
  journal={Advances in neural information processing systems},
  volume={30},
  year={2017}
}

@article{DGCNN,
  title={Dgcnn: A convolutional neural network over large-scale labeled graphs},
  author={Phan, Anh Viet and Le Nguyen, Minh and Nguyen, Yen Lam Hoang and Bui, Lam Thu},
  journal={Neural Networks},
  volume={108},
  year={2018},
  publisher={Elsevier}
}

@inproceedings{PointCLIP,
  title={Pointclip: Point cloud understanding by clip},
  author={Zhang, Renrui and Guo, Ziyu and Zhang, Wei and Li, Kunchang and Miao, Xupeng and Cui, Bin and Qiao, Yu and Gao, Peng and Li, Hongsheng},
  booktitle={Proceedings of the IEEE/CVF conference on computer vision and pattern recognition},
  year={2022}
}

@inproceedings{PointCLIPV2,
  title={Pointclip v2: Prompting clip and gpt for powerful 3d open-world learning},
  author={Zhu, Xiangyang and Zhang, Renrui and He, Bowei and Guo, Ziyu and Zeng, Ziyao and Qin, Zipeng and Zhang, Shanghang and Gao, Peng},
  booktitle={Proceedings of the IEEE/CVF International Conference on Computer Vision},
  year={2023}
}

@inproceedings{CLIP2Point,
  title={Clip2point: Transfer clip to point cloud classification with image-depth pre-training},
  author={Huang, Tianyu and Dong, Bowen and Yang, Yunhan and Huang, Xiaoshui and Lau, Rynson WH and Ouyang, Wanli and Zuo, Wangmeng},
  booktitle={Proceedings of the IEEE/CVF International Conference on Computer Vision},
  year={2023}
}

@inproceedings{ImageNet,
  title={Imagenet: A large-scale hierarchical image database},
  author={Deng, Jia and Dong, Wei and Socher, Richard and Li, Li-Jia and Li, Kai and Fei-Fei, Li},
  booktitle={2009 IEEE conference on computer vision and pattern recognition},
  year={2009},
  organization={Ieee}
}

@article{ShapeNet,
  title={Shapenet: An information-rich 3d model repository},
  author={Chang, Angel X and Funkhouser, Thomas and Guibas, Leonidas and Hanrahan, Pat and Huang, Qixing and Li, Zimo and Savarese, Silvio and Savva, Manolis and Song, Shuran and Su, Hao and others},
  journal={arXiv preprint arXiv:1512.03012},
  year={2015}
}

@inproceedings{ULIP,
  title={Ulip: Learning a unified representation of language, images, and point clouds for 3d understanding},
  author={Xue, Le and Gao, Mingfei and Xing, Chen and Mart{\'\i}n-Mart{\'\i}n, Roberto and Wu, Jiajun and Xiong, Caiming and Xu, Ran and Niebles, Juan Carlos and Savarese, Silvio},
  booktitle={Proceedings of the IEEE/CVF conference on computer vision and pattern recognition},
  year={2023}
}

@article{GPT3,
  title={Language models are few-shot learners},
  author={Brown, Tom and Mann, Benjamin and Ryder, Nick and Subbiah, Melanie and Kaplan, Jared D and Dhariwal, Prafulla and Neelakantan, Arvind and Shyam, Pranav and Sastry, Girish and Askell, Amanda and others},
  journal={Advances in neural information processing systems},
  volume={33},
  year={2020}
}

@article{InfoNCELoss,
  title={Representation learning with contrastive predictive coding},
  author={Oord, Aaron van den and Li, Yazhe and Vinyals, Oriol},
  journal={arXiv preprint arXiv:1807.03748},
  year={2018}
}

@inproceedings{ModelNet,
  title={3d shapenets: A deep representation for volumetric shapes},
  author={Wu, Zhirong and Song, Shuran and Khosla, Aditya and Yu, Fisher and Zhang, Linguang and Tang, Xiaoou and Xiao, Jianxiong},
  booktitle={Proceedings of the IEEE conference on computer vision and pattern recognition},
  year={2015}
}

@inproceedings{dosovitskiy2021an,
title={An Image is Worth 16x16 Words: Transformers for Image Recognition at Scale},
author={Alexey Dosovitskiy and Lucas Beyer and Alexander Kolesnikov and Dirk Weissenborn and Xiaohua Zhai and Thomas Unterthiner and Mostafa Dehghani and Matthias Minderer and Georg Heigold and Sylvain Gelly and Jakob Uszkoreit and Neil Houlsby},
booktitle={International Conference on Learning Representations},
year={2021}
}

@inproceedings{maturana2015voxnet,
  title={Voxnet: A 3d convolutional neural network for real-time object recognition},
  author={Maturana, Daniel and Scherer, Sebastian},
  booktitle={2015 IEEE/RSJ international conference on intelligent robots and systems (IROS)},
  year={2015},
  organization={IEEE}
}

@article{li2018pointcnn,
  title={Pointcnn: Convolution on x-transformed points},
  author={Li, Yangyan and Bu, Rui and Sun, Mingchao and Wu, Wei and Di, Xinhan and Chen, Baoquan},
  journal={Advances in neural information processing systems},
  volume={31},
  year={2018}
}

@article{atzmon2018point,
  title={Point convolutional neural networks by extension operators},
  author={Atzmon, Matan and Maron, Haggai and Lipman, Yaron},
  journal={ACM Transactions on Graphics (TOG)},
  volume={37},
  number={4},
  year={2018},
  publisher={ACM New York, NY, USA}
}

@inproceedings{marethinking,
  title={Rethinking Network Design and Local Geometry in Point Cloud: A Simple Residual MLP Framework},
  author={Ma, Xu and Qin, Can and You, Haoxuan and Ran, Haoxi and Fu, Yun},
  booktitle={International Conference on Learning Representations},
  year={2022}
}
}

\end{document}